\def\year{2018}\relax
\documentclass[letterpaper]{article} 
\usepackage{aaai18}  
\usepackage{times}  
\usepackage{helvet}  
\usepackage{courier}  
\usepackage{url}  
\usepackage{graphicx}  

\usepackage{epsfig}
\usepackage{amsmath}
\usepackage{amssymb}
\usepackage{bm}
\usepackage{romannum}
\usepackage{array}
\usepackage{color}

\begin{document}
%
\title{Building Deep Networks on Grassmann Manifolds}
\author{Zhiwu Huang$^\dagger$, Jiqing Wu$^\dagger$, Luc Van Gool$^{\dagger\ddagger}$\\
	$^\dagger$Computer Vision Lab, ETH Zurich, Switzerland \quad $^\ddagger$VISICS, KU Leuven, Belgium\\
	{\tt\small \{zhiwu.huang, jiqing.wu, vangool\}@vision.ee.ethz.ch}}
\maketitle
\begin{abstract}
Learning representations on Grassmann manifolds is popular in quite a few visual recognition tasks. In order to enable deep learning on Grassmann manifolds, this paper proposes a deep network architecture by generalizing the Euclidean network paradigm to Grassmann manifolds. In particular, we design full rank mapping layers to transform input Grassmannian data to more desirable ones, exploit re-orthonormalization layers to normalize the resulting matrices, study projection pooling layers to reduce the model complexity in the Grassmannian context, and devise projection mapping layers to respect Grassmannian geometry and meanwhile achieve Euclidean forms for regular output layers. To train the Grassmann networks, we exploit a stochastic gradient descent setting on manifolds of the connection weights, and study a matrix generalization of backpropagation to update the structured data. The evaluations on three visual recognition tasks show that our Grassmann networks have clear advantages over existing Grassmann learning methods, and achieve results comparable with state-of-the-art approaches.
\end{abstract}

\section{Introduction}
\label{introduction}

This paper introduces a deep network architecture on Grassmannians, which are manifolds of linear subspaces. In many computer vision applications, linear subspaces have become a core representation. For example, 
for face verification \cite{huang2015projection}, emotion estimation \cite{liu2014combining} and activity recognition \cite{cherian2017generalized}, the image sets of a single person are often modeled by low dimensional subspaces that are then compared on Grassmannians. Besides, for video classification, it is also very common to use autoregressive and moving average (ARMA) model  \cite{vemulapalli2013kernel}. The parameters of the ARMA model are known to be modeled with a high-dimensional linear subspace. For shape analysis, the widely-used affine and linear shape spaces for specific configurations can be also identified by points on the Grassmann manifold \cite{anirudh2017elastic}. Applications involving dynamic environments and autonomous agents often perform online visual learning by using subspace tracking techniques like incremental principal component analysis (PCA) to dynamically learn a better representational model as the appearance of the moving target \cite{turaga2011statistical}.

The popular applications of Grassmannian data motivate us to build a deep neural network architecture for Grassmannian representation learning. To this end, the new network architecture is designed to accept Grassmannian data directly as input, and learns new favorable Grassmannian representations that are able to improve the final visual recognition tasks. In other words, the new network aims to deeply learn Grassmannian features on their underlying Riemannian manifolds in an end-to-end learning architecture. In summary, two main contributions are made by this paper:
\begin{itemize}
	\item We explore a novel deep network architecture in the context of Grassmann manifolds, where it has not been possible to apply deep neural networks. More generally, treating Grassmannian data in deep networks can be very valuable in a variety of machine learning applications.
	
	\item We generalize backpropagation to train the proposed network with deriving an connection weight update rule on a specific Riemannian manifold. Furthermore, we incorporate QR decomposition into backpropagation that might prove very useful in other applications since QR decomposition is a very common linear algebra operator.
\end{itemize}

\section{Background} 

\subsection{Grassmannian Geometry}

A Grassmann manifold $Gr(q,D)$ is a $q(D-q)$ dimensional compact Riemannian manifold, which is the set of $q$-dimensional linear subspaces of the $\mathbb{R}^D$. Thus, each point on $Gr(q,D)$ is a linear subspace that is spanned by the related orthonormal basis matrix $\bm{X}$ of size $D \times q$ such that $\bm{X}^T\bm{X}=\bm{I}_q$, where $\bm{I}_q$ is the identity matrix of size $q \times q$.

One of the most popular approaches to represent linear subspaces and approximate the true Grassmannian geodesic distance is the projection mapping framework $\Phi(\bm{X})=\bm{X}\bm{X}^T$ proposed by \cite{edelman1998geometry}. As the projection $\Phi(\bm{X})$ is a $D \times D$ symmetric matrix, a natural choice of inner product is $\langle \bm{X}_1, \bm{X}_2 \rangle_\Phi=tr(\Phi(\bm{X}_1)^T\Phi(\bm{X}_2))$. The inner product induces a distance metric named projection metric:
\begin{equation}
	\begin{aligned}
		d_p(\bm{X}_1, \bm{X}_2) = 2^{-1/2}\|\bm{X}_1\bm{X}_1^T-\bm{X}_2\bm{X}_2^T\|_F.
	\end{aligned}
	\label{Eq0.7}
\end{equation}
where $\|\cdot\|_F$ indicates the matrix Frobenius norm. As proved in \cite{harandi2013dictionary}, the projection metric can approximate the true geodesic distance up to a scale of $\sqrt{2}$.

\subsection{Grassmann Learning}


To perform discriminant learning on Grassmann manifolds, many works \cite{hamm2008gda,hamm2008extended,cetingul2009intrinsic,harandi2011ggda,harandi2013dictionary,harandi2014expanding} either adopt tangent space approximation of the underlying manifolds, or exploit positive definite kernel functions to embed the manifolds into reproducing kernel Hilbert spaces. In both of such two cases, any existing Euclidean techniques can then be applied to the embedded data, since Hilbert spaces respect Euclidean geometry as well. For example, \cite{hamm2008gda} first embeds the Grassmannian into a high dimensional Hilbert space, and then applies traditional Fisher analysis methods. Obviously, most of these methods are limited to the Mercer \mbox{kernels}, and hence are restricted to use only kernel-based classifiers. Moreover, their computational complexity increases steeply with the growing number of training samples.

More recently, a new learning scheme was proposed by \cite{huang2015projection} to perform a geometry-aware dimensionality reduction from the original Grassmann manifold to another lower-dimensional, more discriminative Grassmann manifold. This could better preserve the original Riemannian data structure, which commonly leads to more favorable classification performances as studied in classical manifold learning. While \cite{huang2015projection} has reached some success, it merely adopts a shallow learning scheme on Grassmann manifolds, which is still far away from the best solution for the problem of representation learning on non-linear manifolds. Accordingly, this paper attempts to open up a possibility of deep learning on Grassmannians.

\begin{figure*}[t]
	\begin{center}
		\includegraphics[width=0.9\linewidth]{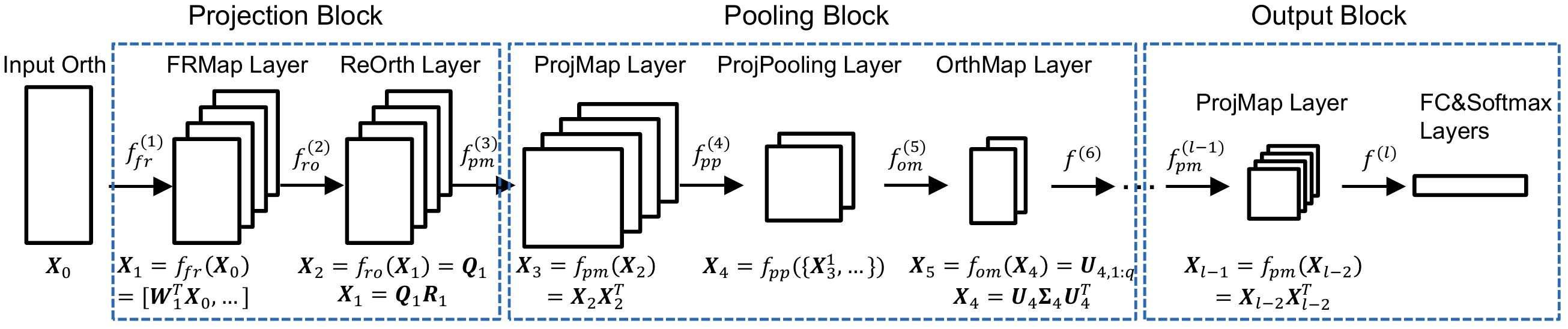}
	\end{center}
	\caption{Conceptual illustration of the proposed Grassmann Network (GrNet) architecture. The rectangles in blue represent three basic blocks, i.e., Projection, Pooling and Output blocks, respectively.}
	\label{fig:long}
	\label{Fig1}
\end{figure*}

\subsection{Manifold Network}

By leveraging the paradigm of traditional neural networks, an increasing number of networks \cite{masci2015geodesic,ionescu2016training,huang2016riemannian} have been built over general manifold domains. For instance, \cite{masci2015geodesic} proposed a `geodesic convolution' on local geodesic coordinate systems to extract local patches on the shape manifold for shape analysis. In particular, the method implements convolutions by sliding a window over the shape manifold, and local geodesic coordinates are employed instead of image patches. In \cite{huang2016riemannian}, to deeply learn appropriate features on the manifolds of symmetric positive definite (SPD) matrices, a deep network structure was developed with some spectral layers, which can be trained by a variant of backpropagation. Nevertheless, to the best of our knowledge, this is the first work that studies a deep network architecture on Grassmann manifolds.

\section{Grassmann Network Architecture}

In analogy to convolutional networks (ConvNets), the proposed Grassmann network (GrNet) also devises a Projection block containing fully connected convolution-like layers and normalization layers, named full rank mapping (FRMap) layers and re-orthonormalization (ReOrth) layers respectively. Inspired by the geometry-aware manifold learning idea \cite{huang2015projection}, the FRMap layers are proposed to firstly perform transformations on input orthonormal matrices of subspaces to generate new matrices by adopting a full rank mapping scheme. Then, the ReOrth layers are developed to normalize the output matrices of the FRMap layers so that they can keep the basic orthogonality. In other words, the normalized data become orthonormal matrices that reside on Stiefel manifold\footnote{A Stiefel manifold $St(d_{k},d_{k-1})$ is the set of $d_{k}$-dimensional orthogonal matrices of the $\mathbb{R}^{d_{k-1}}$.}. As well-studied in \cite{edelman1998geometry,hamm2008gda}, the projection metric performing on orthonormal matrices can represent linear subspaces and respect the geometry of Grassmann manifolds, which is actually a quotient manifold of the Stiefel manifold. Accordingly, we develop projection mapping (ProjMap) layers to maintain the Grassmannian property of the resulting data. Meanwhile, the ProjMap layers are able to transfer the resulting Grassmannian data into Euclidean data, which enables the regular Euclidean layers such as softmax layers for classification. The ProjMap and softmax layers forms the Output block for GrNet. Additionally, since traditional pooling layers can reduce the network complexity, we also study projection pooling (ProjPooling) layers on the projection matrix form of the resulting orthonormal matrices. As it is non-trivial to perform pooling on non-Euclidean data directly, we develop a Pooling block to combine ProMap, ProjPooling and orthonormal mapping (OrthMap) layers, which respectively achieves Euclidean representation, performs pooling on resulting Euclidean data and transforms the results back to orthonormal data. The proposed GrNet structure is illustrated in Fig.\ref{Fig1}.

\subsection{FRMap Layer}

To learn compact and discriminative Grassmannian representation for better classification, we design the FRMap layers to first transform the input orthonormal matrices of subspaces to new matrices by a linear mapping function $f_{fr}$ as
\begin{equation}
	\bm{X}_k = f_{fr}^{(k)}(\bm{X}_{k-1}; \bm{W}_k)=\bm{W}_k\bm{X}_{k-1},
	\label{Eq1}
\end{equation}
where $\bm{X}_{k-1} \in Gr(q,d_{k-1})$ is the input of the $k$-th layer\footnote{For consistency, $k$ is used to denote the relative index of each involved layer in the sequel.}, $\bm{W}_k  \in  \mathbb{R}_{*}^{d_{k} \times d_{k-1}}, (d_{k} < d_{k-1})$ is the transformation matrix (connection weights) that is basically required to be a \mbox{row} full rank matrix, $\bm{X}_k \in \mathbb{R}^{d_{k} \times q}$ is the resulting matrix. Generally, the transformation result $\bm{W}_k\bm{X}_{k-1}$ is not an orthonormal basis matrix. To tackle this problem, we exploit a normalization strategy of QR decomposition in the following ReOrth layer. In addition, as classical deep networks, multiple projections  $\{\bm{W}_{k}^1, \ldots, \bm{W}_{k}^m\}$ per FRMap layer can be applied on each input orthonormal matrix as well, where $m$ is the number of transformation matrices.

As the weight space $\mathbb{R}_{*}^{d_{k} \times d_{k-1}}$ of full rank matrices on each FRMap layer is a non-compact Stiefel manifold where geodesic distance has no closed form. In view of gradient-steepest-descent learning by geodesic stepping for criterion optimization, it is necessary to have a closed form of the geodesic distance to derive natural gradients over a smooth manifold.  Hence, optimizing on the non-compact manifold directly is infeasible unless endowing the manifold with pseudo-Riemannian metrics. To handle this problem, one feasible solution is imposing orthogonality constraint on the weight matrix $\bm{W}_k$ so that it resides on a compact Stiefel manifold $St(d_{k},d_{k-1})$. Obviously, such orthogonal solution space is smaller than the original solution space $\mathbb{R}^{n \times m}_{*}$, making the optimization theoretically yield suboptimal solution of $\bm{W}_k$. To achieve a more faithful solution, following \cite{huang2015projection}, we alternatively do the optimization over the manifolds $PSD(d_{k},d_{k-1})$\footnote{A PSD manifold $PSD(d_{k},d_{k-1})$ is the set of $d_{k}$-rank positive semidefinite matrices of size ${d_{k-1}}$ \cite{bonnabel2009riemannian,journee2010lowrank}.} of the conjugates $\bm{P}_k=\bm{W}_k^T\bm{W}_k$, which are actually positive semidefinite (PSD) matrices. As studied in \cite{bonnabel2009riemannian,journee2010lowrank,meyer2011regression} and the popular manopt
toolbox\footnote{\url{http://www.manopt.org/}}, a PSD manifold is actually a quotient space, and thus optimizing on it actually pursues optimal full rank matrix $\bm{W}_k$ directly. In the sense, optimizing on PSD manifolds actually minimizes $f(\bm{W}_k)$ instead of $f(\bm{W}_k^T\bm{W}_k)$, where $f$ denotes the involved layer's loss function that will be introduced in the next section.

\subsection{ReOrth Layer}

Inspired by \cite{kim2007discriminative,huang2015projection} that use QR decomposition to transform a regular matrix to an orthonormal matrix, we design the ReOrth layers to perform QR decomposition on the input matrix $\bm{X}_{k-1}$
\begin{equation}
	\bm{X}_{k-1}=\bm{Q}_{k-1}\bm{R}_{k-1},
	\label{Eq1.00}
\end{equation}
where $\bm{Q}_{k-1} \in \mathbb{R}^{d_{k-1} \times q}$ is the orthonormal matrix composed by the first $q$ columns, and $\bm{R}_{k-1} \in \mathbb{R}^{q \times q}$ is the invertible upper-triangular matrix. Since $\bm{R}$ is invertible and $\bm{Q}$ is orthonormal, we can make $\bm{X}_k$ become an orthonormal basis matrix by normalizing $\bm{X}_{k-1}$ in the $k$-th layer as:
\begin{equation}
	\bm{X}_{k}= f_{ro}^{(k)}(\bm{X}_{k-1}) = \bm{X}_{k-1}\bm{R}_{k-1}^{-1} = \bm{Q}_{k-1}.
	\label{Eq1.0}
\end{equation}

In the context of ConvNets, \cite{cybenko1989approximation,jarrett2009best,nair2010rectified,goodfellow2013maxout,he2015delving} have presented various nonlinear activation functions, e.g., rectified linear units (ReLU), to improve discriminative performance. Accordingly, exploiting this kind of layers to introduce the non-linearity to the domain of the proposed GrNet is also necessary. In the light of this, to some extent, the function Eqn.\ref{Eq1.0} also takes a role of performing a non-linear activation with the QR factorization.

\subsection{ProjMap Layer}

The ProjMap layer is designed to perform Grassmannian computation on the resulting orthonormal matrices. As well-studied in \cite{edelman1998geometry,hamm2008gda,hamm2008extended,harandi2011ggda,huang2015projection}, the projection metric is one of the most popular Grassmannian metrics, and is able to endow the specific Riemannian manifold with an inner product structure so that the manifold is reduced to a flat space. In the flat \mbox{space}, classical Euclidean computations can be applied to the projection domain of orthonormal matrices. Formally, we apply the projection mapping \cite{edelman1998geometry} to a orthonormal matrix $\bm{X}_{k-1}$ in the $k$-th layer as
\begin{equation}
	\bm{X}_{k} = f_{pm}^{(k)}(\bm{X}_{k-1})=\bm{X}_{k-1}\bm{X}_{k-1}^T.
	\label{Eq4}
\end{equation}

As for other Riemannian computations on Grassmann manifolds, please refer to \cite{le1991onge,edelman1998geometry,srivastava2004bayesian,absil2008optimization,helmke2007newton}.

\subsection{ProjPooling Layer}

It is known that classical pooling layers with max, min and mean pooling functions reduce the sizes of the representations to lower the model complexity, and therefore improve the regular ConvNets. Motivated by this, we also design pooling layers for the feature maps of Grassmannian data.

Without loss of generality, we here study mean pooling for the Grassmannian points. Actually, there exist some approaches \cite{absil2004riemannian,dodge1999multivariate,srivastava2002monte,marrinan2014finding} to compute mean points on Grassmannians. Inspired by the idea \cite{srivastava2002monte} with keeping the balance between computational time and calculation accuracy, we propose three layers to implement mean pooling for Grassmannian data. In particular, the Grassmannian data are first mapped to the space of projection matrices by the ProjMap layer presented before. As the resulting $m$ projection matrices $\{\bm{X}_{k-1}^i| 1 \leq i \leq m\}$ of size $d_{k-1} \times d_{k-1}$ are Euclidean, we then design a regular mean pooling layer for them. Lastly, we devise an orthonormal mapping (OrthMap) layer to transform the mean projection matrix back to orthonormal data. Formally, the functions for the ProjPooling and OrthMap layers are respectively defined as
\begin{alignat}{2}
	&\bm{X}_{k}  = f_{pp}^{(k)}(\{\hat{\bm{X}}_{k-1}^1, \ldots, \hat{\bm{X}}_{k-1}^n\})= \frac{1}{n}\sum_i^n \hat{\bm{X}}_{k-1}^i, \label{Eq4.0} \\
	&\bm{X}_{k+1}  = f_{om}^{(k)}(\bm{X}_{k})=  \bm{U}_{k-1, 1:q}. \label{Eq4.1}
\end{alignat}
where $\bm{U}_{k-1, 1:q}$ is the first $q$ largest eigenvectors achieved by eigenvalue (EIG) decomposition on the input projection matrices $\bm{X}_{k}=\bm{U}_{k-1}\bm{\Sigma}_{k-1} \bm{U}_{k-1}^T$, and $n$ is the number of instances $\hat{\bm{X}}_{k-1}^i$ for the pooling. The instances can be either $n$ projection matrices or $n$ entries within an square patch of size $\sqrt{n} \times \sqrt{n}$ located in one projection matrix. In other words, the first type of pooling is performed across the projection matrices (A-ProjPooling), while the second one is executed by sliding the mean filter over each square patch within one projection matrix (W-ProjPooling). As a result, A-ProjPooling with OrthMap finally yields $\frac{m}{n}$ orthonormal matrices of size $d_{k-1} \times q$, while W-ProjPooling with OrthMap outputs $m$ orthonormal matrices of size  $\frac{d_{k-1}}{\sqrt{n}} \times q$.

\subsection{Output Layers}

As shown in Fig.\ref{Fig1}, after applying the ProjMap layer, the outputs (i.e., the projection matrices) lie in Euclidean space, and thus can be converted into vector forms. Hence, on the top of the ProjMap layer, classical Euclidean network layers can be employed. For instance, the regular fully connected (FC) layer could be used after the ProjMap layer. The dimensionality of the filters in the FC layer is typically set to $d_{k} \times d_{k-1}$, where $d_{k}$ and $d_{k-1}$ are the class number and the dimensionality of the input vector forms respectively. Finally, the common softmax layer can be used for visual classification.

\section{Training Grassmann Network}
\label{backprop}

As most layers in the GrNet model are expressed with complex matrix factorization functions, they cannot be simply reduced to a constructed bottom-up from element-wise calculations. In other words, the matrix backpropgation (backprop) cannot be derived by using traditional matrix that treats element-wise operations in matrix form. As a result, simply using the traditional backprop will break down in the setting. To solve the problem,  \cite{huang2016riemannian,ionescu2016training} introduced manifold-valued connection weight update rule and matrix backprop respectively. Furthermore, the convergence of the stochastic gradient descent (SGD) algorithm on Riemannian manifolds has also been studied well in \cite{bottou2010large,bonnabel2013stochastic}. Accordingly, we exploit the training procedure for the proposed GrNet upon these existing works.

To begin with, we represent the proposed GrNet model with a sequence of successive function compositions $f= f^{(l)} \circ f^{(l-1)} \circ f^{(l-2)} \ldots \circ f^{(2)} \circ f^{(1)}$ with a parameter tuple $\bm{W} = (\bm{W}_l, \bm{W}_{l-1}, \ldots, \bm{W}_1)$, where $f^{(k)}$ and $\bm{W}_k$ are the function and the weight matrix respectively for the $k$-th layer, and $l$ is the number of layers. The loss of the $k$-th layer can be denoted by a function as $L^{(k)}=\ell \circ f^{(l)} \circ \ldots f^{(k)}$, where $\ell$ is the loss function for the last output layer.

Then, we recall the definition of the matrix backprop and its properties studied in \cite{ionescu2016training}. In particular, \cite{ionescu2016training} exploits a function $\mathcal{F}$ to describe the variations of the upper layer variables with respect to the lower layer variables, i.e., $d\bm{X}_{k}=\mathcal{F}(d\bm{X}_{k-1})$. Consequently, a new version of the chain rule for the matrix backprop is defined as
\begin{equation}
	\frac{\partial L^{(k)}(\bm{X}_{k-1}, y)}{\partial \bm{X}_{k-1}}  = \mathcal{F}^{*} \left(\frac{\partial L^{(k+1)}(\bm{X}_k, y)}{\partial \bm{X}_k}\right),
	\label{Eq10}
\end{equation}
where $y$ is the desired output, $\bm{X}_k=f^{(k)}(\bm{X}_{k-1})$, $\mathcal{F}^{*}$ is a non-linear adjoint operator of $\mathcal{F}$, i.e., $a: \mathcal{F}(b)=\mathcal{F}^{*}(a) : b$, the matrix inner product $\bm{A}:\bm{B} = Tr(\bm{A}^T\bm{B})$.

In the sequel, we will detail the connection weight update on the specific PSD manifold and the matrix backprop process through some key layers in the context of the proposed GrNet. For simplicity, we uniformly let $\partial L^{(k)}(\bm{X}_{k-1}, y)$ be $\partial L^{(k)}$, $\bm{Q}_{k-1}$ be $\bm{Q}$, and $\bm{R}_{k-1}$ be $\bm{R}$ respectively.

\subsection{FRMap Layer}

For the FRMap layers, we propose a new way of updating the weights appeared in Eqn.\ref{Eq1} by exploiting an SGD setting on PSD manifolds. As studied in \cite{absil2008optimization}, the steepest descent direction for the corresponding loss function $L^{(k)}(\bm{X}_{k-1}, y)$ with respect to $\bm{W}_k$ on one Riemannian manifold is the Riemannian gradient $\tilde{\nabla} L^{(k)}_{\bm{W}_k}$. In particular, following the standard optimization \cite{absil2008optimization} on Riemannian manifolds, we first apply the parallel transport to transfer the Euclidean gradient in the tangent space at the current status of the weight $\bm{W}_k^{t}$ to the one in the tangent space at the next status $\bm{W}_{k}^{t+1}$. Then the resulting Euclidean gradient is subtracted to the normal component of the Euclidean gradient $\nabla L^{(k)}_{\bm{W}_k^{t}}$. After this operation, searching along the tangential direction yields the update in the tangent space of the PSD manifold. Finally, the resulting update is mapped back to the PSD manifold with a retraction operation $\Gamma$. For more details about the geometry of PSD manifolds and the retraction operation on Riemannian manifolds, the readers are referred to \cite{bonnabel2009riemannian,journee2010lowrank,meyer2011regression,absil2008optimization}. Accordingly, the update of the current connection weight $\bm{W}_k^{t}$ on the PSD manifold adheres to the following form
\begin{alignat}{2}
	\tilde{\nabla} L^{(k)}_{\bm{W}_k^{t}}&=\nabla L^{(k)}_{\bm{W}_k^{t}}-\nabla L^{(k)}_{\bm{W}_k^{t}}{(\bm{W}_k^{t})}^T\bm{W}_k^{t}, \label{Eq7} \\
	\bm{W}_k^{t+1} &= \Gamma(\bm{W}_k^{t}-\lambda\tilde{\nabla} L^{(k)}_{\bm{W}_k^{t}}), \label{Eq8}
\end{alignat}
where $\lambda$ is the learning rate, $\nabla L^{(k)}_{\bm{W}_k^{t}}{(\bm{W}_k^{t})}^T\bm{W}_k^{t}$ is the normal component of the Euclidean gradient $\nabla L^{(k)}_{\bm{W}_k^{t}}$. By employing the conventional backprop, $\nabla L^{(k)}_{\bm{W}_k^{t}}$ is computed by
\begin{equation}
	\nabla L^{(k)}_{\bm{W}_k^{t}}=\frac{\partial L^{(k+1)}}{\partial \bm{X}_k}\frac{ \partial f^{(k)}(\bm{X}_{k-1})}{\partial \bm{W}_k^{t}}=\frac{\partial L^{(k+1)}}{\partial \bm{X}_k}\bm{X}_{k-1}^T.
	\label{Eq9}
\end{equation}

\subsection{ReOrth Layer}

Actually, the ReOrth layers involve QR decomposition Eqn.\ref{Eq1.00} and the non-linear operation Eqn.\ref{Eq1.0}. Firstly, for Eqn.\ref{Eq1.00} we introduce a virtual layer $k^{'}$, which receives $\bm{X}_{k-1}$ as input and produces a tuple $\bm{X}_{k^{'}}=(\bm{Q}$, $\bm{R})$. Following \cite{ionescu2016training} to handle the case of a tuple output, we apply the new chain rule Eqn.\ref{Eq10} with the equations $a: \mathcal{F}(b)=\mathcal{F}^{*}(a) : b$ and $d\bm{X}_{k^{'}}=\mathcal{F}(d\bm{X}_{k-1})$ to achieve the update rule for the structured data:
\begin{equation}
	\begin{aligned}
		& \frac{\partial L^{(k)}}{\partial \bm{X}_{k-1}}: d\bm{X}_{k-1} \\
		& =  \mathcal{F}^{*} \left(\frac{\partial L^{(k^{'})}}{\partial \bm{Q}}\right): d\bm{X}_{k-1} + \mathcal{F}^{*} \left(\frac{\partial L^{(k^{'})}}{\partial \bm{R}}\right): d\bm{X}_{k-1} \\
		&= \frac{\partial L^{(k^{'})}}{\partial \bm{Q}} : \mathcal{F} \left(d\bm{X}_{k-1}\right) + \frac{\partial L^{(k^{'})}}{\partial \bm{R}} : \mathcal{F} \left(d\bm{X}_{k-1}\right) \\
		&= \frac{\partial L^{(k^{'})}}{\partial \bm{Q}} : d\bm{Q}+ \frac{\partial L^{(k^{'})}}{\partial \bm{R}} : d\bm{R},
	\end{aligned}
	\label{Eq11}
\end{equation}
where the two variations $d\bm{Q}$ and $d\bm{R}$ are derived by the variation of the QR operation $d\bm{X}_{k-1}=d\bm{Q}\bm{R}+ \bm{Q}d\bm{R}$ as:
\begin{alignat}{2}
	d\bm{Q} & =\bm{S}d\bm{X}_{k-1}\bm{R}^{-1}+\bm{Q}(\bm{Q}^T d\bm{X}_{k-1} \bm{R}^{-1})_{asym}, \label{Eq12} \\
	d\bm{R} &=\bm{Q}^T d\bm{X}_{k-1}-(\bm{Q}^T d\bm{X}_{k-1} \bm{R}^{-1})_{asym}\bm{R}, \label{Eq13}
\end{alignat}
where $\bm{S}=\bm{I}-\bm{QQ}^T$, $\bm{I}$ is an identity matrix, $\bm{A}_{asym}=\bm{A}_{tril}-(\bm{A}_{tril})^T$, $\bm{A}_{tril}$ extracts the elements below the main diagonal of $\bm{A}$. For more details to derive Eqn.\ref{Eq12} and Eqn.\ref{Eq13}, please refer to the part \Romannum{1} of Appendix.

As derived by the part \Romannum{2} of Appendix, plugging Eqn.\ref{Eq12} and Eqn.\ref{Eq13} into Eqn.\ref{Eq11} achieves the partial derivatives of the loss functions for the ReOrth layers:
\begin{equation}
	\begin{aligned}
		\frac{\partial L^{(k)}}{\partial \bm{X}_{k-1}} & = \left (\bm{S}^T\frac{\partial L^{(k^{'})}}{\partial \bm{Q}}+\bm{Q}\left(\bm{Q}^T \frac{\partial L^{(k^{'})}}{\partial \bm{Q}}\right)_{bsym}\right) {(\bm{R}^{-1})}^T\\
		& + \bm{Q} \left( \frac{\partial L^{(k^{'})}}{\partial \bm{R}}-\left(\frac{\partial L^{(k^{'})}}{\partial \bm{R}} \bm{R}^{T}\right)_{bsym}(\bm{R}^{-1})^T\right),
		\label{Eq15}
	\end{aligned}
\end{equation}
where $\bm{A}_{bsym}=\bm{A}_{tril}-(\bm{A}^T)_{tril}$, $\bm{A}_{tril}$ extracts the elements below the main diagonal of $\bm{A}$. $\frac{\partial L^{(k^{'})}}{\partial \bm{Q}}$ and $\frac{\partial L^{(k^{'})}}{\partial \bm{R}}$ can then be obtained on the function Eqn.\ref{Eq1.0} employed in the ReOrth layers. Specially, its variation becomes $d\bm{X}_{k}=d\bm{Q}$.
Therefore, the involved partial derivatives with respect to $\bm{Q}$ and $\bm{R}$ are computed by $\frac{\partial L^{(k^{'})}}{\partial \bm{Q}} = \frac{\partial L^{(k+1)}}{\partial \bm{X}_k}$ and $\frac{\partial L^{(k^{'})}}{\partial \bm{R}} = 0$.


\subsection{OrthMap Layer}

As presented before, the OrthMap layers involve eigenvalue (EIG) decomposition. Thus we adopt the proposition in \cite{ionescu2016training} to calculate the partial derivatives for the EIG computation.


\setlength{\parskip}{1\baselineskip}

\noindent \textbf{Proposition 1} \emph{Let $\bm{X}_{k-1}=\bm{U\Sigma U}^T$ with  $\bm{X}_{k-1} \in \mathbb{R}^{D\times D}$, such that $\bm{U}^T\bm{U} = \bm{I}$ and $\bm{\Sigma}$ owns a diagonal structure. The resulting partial derivative for the EIG layer $k^{'}$ is given by}
\begin{equation}
	\begin{aligned}
		\frac{\partial L^{(k)}}{\partial \bm{X}_{k-1}} &= \bm{U} \left(\hat{\bm{K}}^T \circ \left(\bm{U}^T \frac{\partial L^{(k^{'})}}{\partial \bm{U}} \right) \right)\bm{U}^T \\
		& +  \bm{U}\left(\frac{\partial L^{(k^{'})}}{\partial \bm{\Sigma}}\right)_{diag} \bm{U}^T.
		\label{Eq17.0}
	\end{aligned}
\end{equation}
where $\hat{\bm{K}}=1/(\sigma_i-\sigma_j), i \neq j; 0, i=j$ ($\sigma_i$ is the diagonal element of $\bm{\Sigma}$), and the partial derivatives with respect to $\bm{\Sigma}$ and $\bm{U}$ in Eqn.\ref{Eq4.1} for the OrthMap layers can be achieved by $\frac{\partial L^{(k^{'})}}{\partial \bm{U}}  = [\frac{\partial L^{(k+1)}}{\partial \bm{X}_k} \quad \bm{0}]$ and $\frac{\partial L^{(k^{'})}}{\partial \bm{\Sigma}} = 0$,
where $\bm{0}$ is the matrix of size $D \times (D-q)$ with all elements being zero.

\section{Empirical Evaluation}

We compare four groups of exiting methods to evaluate the proposed GrNet for three visual classification tasks: emotion recognition, action recognition and face verification.

\noindent\textbf{Comparing methods}: \textbf{1).} \emph{General manifold learning methods:} 
  Expressionlets on Spatio-Temporal Manifold (STM-ExpLet) \cite{liu2014learning} and Riemannian Sparse Representaion combining with Manifold Learning on the manifold of SPD matrices (RSR-SPDML) \cite{harandi2014manifold};
 \textbf{2).} \emph{Grassmann learning methods}:
   Discriminative Canonical Correlations (DCC) \cite{kim2007discriminative}, Grassmann Discriminant Analysis (GDA) \cite{hamm2008gda}, Grassmannian Graph-Embedding Discriminant Analysis (GGDA) \cite{hamm2008extended} and Projection Metric Learning (PML) \cite{huang2015projection};
  \textbf{3).} \emph{Regular convolutional networks}:
  VGGDeepFace \cite{parkhi2015deep}
  and Deep Second-order Pooling (DeepO2P) \cite{ionescu2016training};
  \textbf{4).} \emph{Manifold network}:
  Network on SPD manifolds (SPDNet) \cite{huang2016riemannian} and DeepO2P that trains standard ConvNets with a manifold layer end-to-end

We use source codes of all the comparing methods from authors with tuning their parameters as in the original papers. For our GrNet, we build its architecture with using $i$ Projection-Pooling block(s) (named as GrNet-$i$Block) and one Output block, all of which are illustrated in Fig.\ref{Fig1}. The learning rate $\lambda$ and the batch size are set to $0.01$ and 30 respectively. The FRMap matrices are all initialized as random full rank matrices, and the number of them per layer is set to 16. For all the ProjPooling layers, the number $n$ of the instances for pooling are fixed as 4. For training the GrNet, we just use an i7-2600K (3.40GHz) PC without any GPUs\footnote{As the matrix factorizations are implemented well in CUDA, we will achieve the GPU version of our GrNet for speedups.}. Note that, the readers can follow the DeepO2P method to implement an end-to-end learning of ConvNet+GrNet. Since the focus of this paper is on deep learning for Grassmannian inputs, we leave the study in future work.

\noindent\textbf{Emotion Recognition}: We utilize the popular Acted Facial Expression in Wild (AFEW) \cite{dhall2014emotion} dataset. The dataset contains 1,345 sequences of facial expressions acted by 330 actors in close to real world setting. The standard protocol designed by \cite{dhall2014emotion} splits the dataset into three data sets, i.e., training, validation and test data sets. In the training and validation data sets, each video is classified into one of seven expressions, while the ground truth of the test set has not been released. As a result, we follow \cite{liu2014learning,huang2016riemannian} to present the results on the validation set.
As done in many works such as \cite{huang2016riemannian} for augmenting the training data, we split the training videos to 1,747 small subvideos. For the evaluation, each facial frame is normalized to an image of size $20 \times 20$. Then, following \cite{liu2013partial,liu2014combining}, we model sequences of facial expression with a set of linear subspaces of order 10, which span a Grassmann manifold $Gr(10,400)$. In the task, the sizes of the GrNet-1Block weights are set to $400 \times 100$, while those of the GrNet-2Blocks are set to $400 \times 300$ and $150 \times 100$.

\noindent\textbf{Action Recognition}: We use the HDM05 database \cite{muller2007hdm} that is one of the largest-scale skeleton-based human action datasets. The dataset consists of 2,337 sequences of 130 action classes, and provides 3D locations of 31 joints of the subjects.
Following the protocol designed in \cite{huang2016riemannian}, we conduct 10 random evaluations, each of which randomly selected half of sequences for training and the rest for testing. For data augmentation, the training sequences are divided into around 18,000 small subsequences in each random evaluation.
As done in \cite{harandi2014manifold,huang2016riemannian}, we represent each sequence by a covariance descriptor of size $93 \times 93$, which is computed by the second order statistics of the 3D coordinates of the 31 joints in each frame. Then, we apply SVD on the covariance descriptors to get linear subspaces of order 10, which form the data on a Grassmannian $Gr(10,93)$. For our GrNet-1Block, the sizes of the connection weights are set to  $93 \times 60$, while those of GrNet-2Blocks are fixed as $93 \times 80, 40 \times 30$ respectively.

\noindent\textbf{Face Verification}: We employ one standard dataset named Point-and-Shoot Challenge (PaSC) \cite{beveridge2013challenge}. For 256 subjects, it owns 1,401 videos taken by control cameras, and 1,401 videos from handheld cameras.
For the dataset, \cite{beveridge2013challenge} designs control and handheld face verification experiments. As done in \cite{beveridge2015report,huang2016riemannian}, we use its 280 training videos and the COX data \cite{huang2015benchmark} with 900 videos for training. Similarly, the training data are split to 12,529 small clips. To extract the state-of-the-art deep features, we perform the approach of \cite{parkhi2015deep} on the normalized face images of size $224 \times 224$. To speed up the training, we employ PCA to reduce the deep features to 400-dimensional ones. Following \cite{huang2016riemannian}, a SPD matrix of size $401 \times 401$ is computed by fusing covariance matrix and mean for each video. As done in \cite{huang2015projection} on each video, we finally compute a linear subspace of order 10, which lies on $Gr(10,401)$. We set the sizes of GrNet-1Block weights to $401 \times 100$, while setting those to $401 \times 300$ and $150 \times 100$ for GrNet-2Blocks.

\noindent\textbf{{Experimental Analysis}}

Table.\ref{tab1} presents the performances of the comparing methods for the three used datasets. 
The results show our GrNet with 2 blocks can outperform the existing general manifold learning, Grassmann learning methods and standard ConvNets (i.e., VGGDeepFace and DeepO2P). Particularly, on HDM05, we can observe that our GrNet outperforms the state-of-the-art Grassmann learning methods by a large margin (more than 11\%). This verifies that the proposed GrNet yields great improvements when the training data is large enough. For PaSC, although the used softmax output layer in the GrNet does not suit the verification task well, we find that it still reaches the highest performances in the case of 2 blocks, which learns more favorable Grassmannian representation. As studied in existing state-of-the-art Grassmann learning methods, the GrNet without Riemannian computing (i.e., ProjMap) and without geometry-aware learning (i.e., ReOrth) perform very badly (17.62\% and 26.15\% respectively for AFEW). Besides, the consistent improvement of stacking more GrNet blocks verifies it can learn more discriminative Grassmannian representations and finally improve the classification performances.

By comparing one of the manifold networks DeepO2P that uses an end-to-end training of ConvNet with a single SPD manifold layer, our GrNets achieve better performances. In contrast, our GrNets fail to surpass the other manifold network SPDNets that use multiple SPD manifold layers. Nevertheless, our GrNets perform deep learning on a different type of manifolds with owning many considerable differences in both terms of application range and intrinsic properties, some of which are enumerated below.

\begin{figure}
	\begin{center}
		\includegraphics[width=0.9\linewidth]{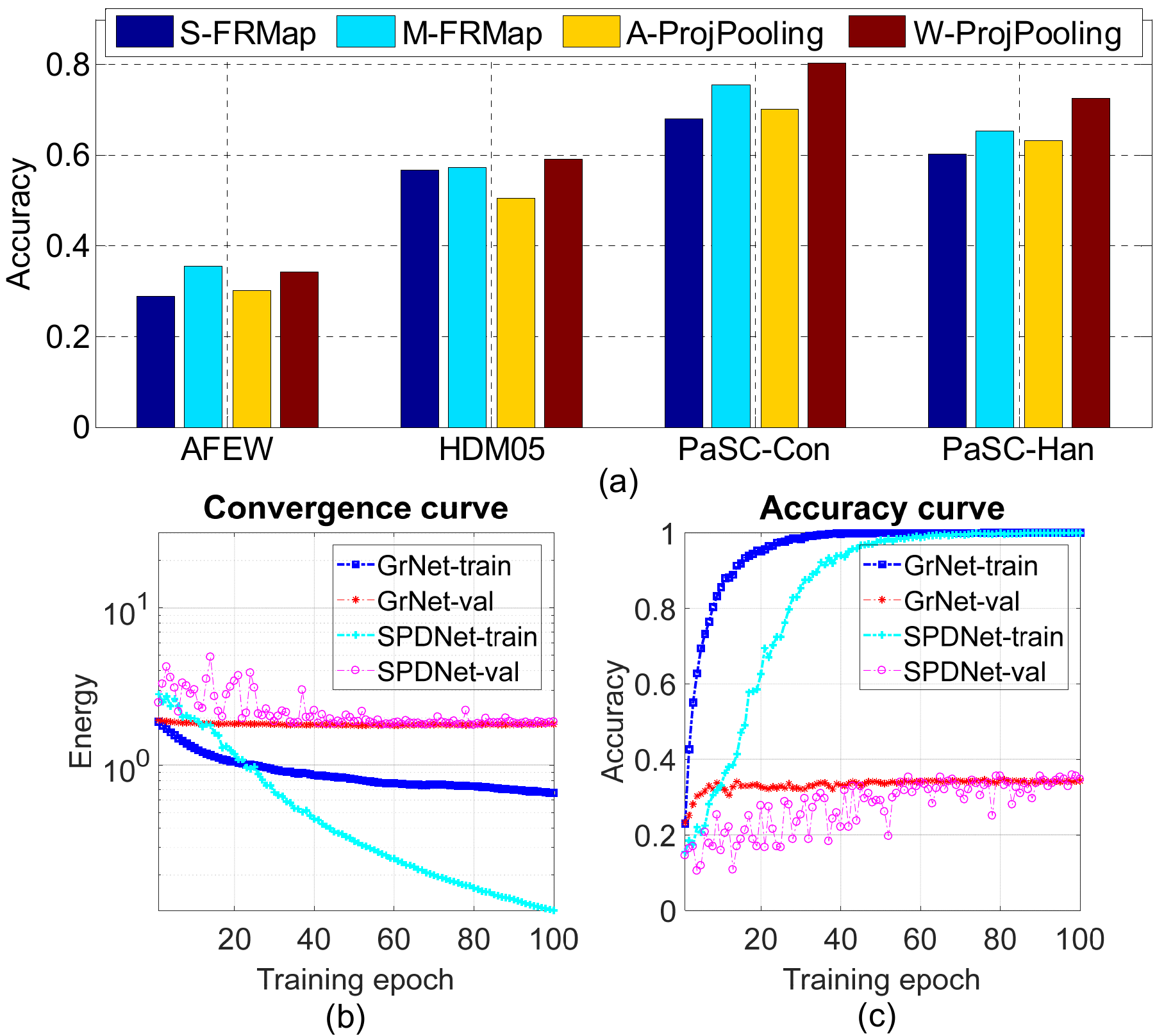}
		\caption{(a) Results of using single and multiple FRMap (S-FRMap, M-FRMap), ProjPoolings across or within projections (A-ProjPooling, W-ProjPooling) for the three used databases. (b) (c) Convergence and accuracy curves of SPDNet and  the proposed GrNet for the AFEW. }
		\label{Fig3}
	\end{center}
\end{figure}

1). The proposed FPMap layers learn full rank projections while the BiMap layers in SPDNets pursue bi-linear orthogonal projections. Besides, the GrNets exploit both single and multiple projections in each FRMap layer (i.e., S-FRMap, M-FRMap). The results in Fig.\ref{Fig3} (a) show the benefit of using M-FRMap. Furthermore, it alternatively studies a new connection weight update rule on PSD manifolds rather than the one on Stiefel manifolds, whose performances (i.e., 32.13\%, 57.25\%, 80.15\%, 71.28\%) on the datasets are often worse.

2). The GrNets design brand new ReOrth, ProjMap and ProjPooling layers. Particularly, for the ReOrth layers, the exploration of the QR decomposition in backpropagation is a very important theoretical contribution. Besides, it devises pooling layer across and within projection matrices (A-ProjPooling, W-ProjPooling), both of which work on the top of the M-FRMap layers. As studied in Fig.\ref{Fig3} (a), W-ProjPooling typically outperforms A-ProjPooling. The unfavorable performance of A-ProjPoolings would be caused by the extrinsic mean calulation on the Grassmannian and the weak relationship across multiple projection matrices.

3). Training GrNet-1Block per epoch costs about 10, 9 and 13 minutes respectively on the three datasets, while training SPDNet (w/o complex pooling) takes 2, 4 and 15 minutes. However, in theory, our GrNet actually runs much faster than the existing SPDNet when using the same setting. This is because the GrNet handles much lower-dimensional orthonormal matrices of size $d \times q$ (the order $q$ is often set to 10), while the SPDNet treats SPD matrices of size $d \times d$. 

4). Fig.\ref{Fig3} (b)(c) show the GrNet can use much less epochs (than the SPDNet) to converge on AFEW, and the validation gets near 12\% improvement after training. For larger datasets like HDM05, the improvement is even up to 40\%.

\begin{table}
	\small
	\begin{center}
		\begin{tabular}{|l|m{0.85cm}<{\centering}|m{1.65cm}<{\centering}|m{0.85cm}<{\centering}m{0.85cm}<{\centering}|}
			\hline
			Method & AFEW & HDM05 & PaSC1 & PaSC2\\
			\hline\hline
	        STM-ExpLet & 31.73\%  & --  & --  & --\\
			RSR-SPDML & 30.12\% & 48.01\%$\pm$3.38  & --  & --\\
			
			\hline
			
			DCC & 25.78\% & 41.34\%$\pm$1.05 & 75.83\% & 67.04\% \\
			GDA & 29.11\%  & 46.25\%$\pm$2.71 & 71.38\% & 67.49\% \\
			GGDA  & 29.45\%   & 46.87\%$\pm$2.31  & 66.71\% & 68.41\%  \\
			PML & 28.98\% & 47.12\%$\pm$1.59 & 73.45\% & 68.32\%   \\
			\hline

			VGGDeepFace & --  & --  & 78.82\% & 68.24\%\\
			DeepO2P & 28.54\% & -- & 68.76\% & 60.14\%\\
						
			\hline
			SPDNet & 34.23\% & 61.45\%$\pm$1.12 & 80.12\% & 72.83\% \\

			\hline
			GrNet-0Block & 25.34\% & 51.12\%$\pm$3.55  & 68.52\% & 63.92\% \\
			GrNet-1Block & 32.08\% & 57.73\%$\pm$2.24 &  80.15\% & 72.51\% \\
			GrNet-2Blocks & 34.23\% & 59.23\%$\pm$1.78 & 80.52\% & 72.76\% \\
			\hline
		\end{tabular}
	\end{center}
	\caption{Results for the AFEW, HDM05 and PaSC datasets. PaSC1/PaSC2 are the control/handheld testings.}
	\label{tab1}
\end{table}

\section{Conclusion}

This paper introduced the first network architecture to perform deep learning over Grassmann manifolds. Essentially, it is a natural exploration of convolutional networks to perform fully connected convolution, normalization, pooling and Remannian computing on Grassmannian data. In three typical visual classification evaluations, our Grassmann networks significantly outperformed existing Grassmann learning methods, and performed comparably with state-of-the-art methods.  
Directions for future work include extending the current GrNet to an end-to-end ConvNet+GrNet training system and applying it to other computer vision problems.
\textbf{Acknowledgement:} This work is supported by EU Framework Seven project ReMeDi (grant 610902).

\section*{Appendix}

\textbf{\Romannum{1}}. Regarding the gradient computation of QR decomposition, we first differentiate its implicit system
\begin{alignat}{1}
	\bm{X}_{k-1}  = \bm{QR}, \quad
	\bm{Q}^T\bm{Q}  = \bm{I}, \quad
	0  = \bm{R}_{tril},
\end{alignat}
where $\bm{R}_{tril}$ returns the elements below the main diagonal of $\bm{R}$, and we obtain
\begin{alignat}{1}
	d\bm{X}_{k-1} =  d\bm{QR}+\bm{Q}d\bm{R}, \quad
	d\bm{Q}^T\bm{Q} = -\bm{Q}^Td\bm{Q}.
	\label{Eq20}
\end{alignat}
Multiplying the first equation of Eqn.\ref{Eq20} from the left with $\bm{Q}^T$ and the right with $\bm{R}^{-1}$ derives
\begin{alignat}{2}
	\quad d\bm{R} & = \bm{Q}^Td\bm{X}_{k-1} - \bm{Q}^Td\bm{QR}, \label{Eq22} \\
	\quad d\bm{Q} & = d\bm{X}\bm{R}^{-1} -\bm{Q}d\bm{R}\bm{R}^{-1}. \label{Eq23.1}
\end{alignat}

The multiplication of Eqn.\ref{Eq22} from the right with the inverse of $\bm{R}$ yields the equation
\begin{alignat}{1}
	0  = \bm{Q}^Td\bm{X}_{k-1}\bm{R}^{-1} - \bm{Q}^Td\bm{QR}\bm{R}^{-1}-d\bm{R}\bm{R}^{-1}. \label{Eq23}
\end{alignat}

As $(d\bm{R}\bm{R}^{-1})_{tril} = 0$, we further derive
$	(\bm{Q}^Td\bm{Q})_{tril}  = (\bm{Q}^Td\bm{X}_{k-1}\bm{R}^{-1})_{tril}. \label{Eq24}$
Since $\bm{Q}^Td\bm{Q}=(\bm{Q}^Td\bm{Q})_{tril}-((\bm{Q}^Td\bm{Q})_{tril})^T $ is antisymmetric (see Eq.\ref{Eq20}) we have
\begin{equation}
	\begin{aligned}
		\bm{Q}^Td\bm{Q}&=(\bm{Q}^Td\bm{X}_{k-1}\bm{R}^{-1})_{tril}-((\bm{Q}^Td\bm{X}_{k-1}\bm{R}^{-1})_{tril})^T \\
		&=(\bm{Q}^Td\bm{X}_{k-1}\bm{R}^{-1})_{asym}.
	\end{aligned}
	\label{Eq25}
\end{equation}

Substituting Eqn.\ref{Eq25} into Eqn.\ref{Eq22} derives the gradient of QR decomposition w.r.t $\bm{R}$ as Eqn.\ref{Eq13}. By plugging Eqn.\ref{Eq13} into Eqn.\ref{Eq23.1}, we can derive the gradient of the QR decomposition w.r.t $\bm{Q}$ as Eqn.\ref{Eq12}.



%
%

\textbf{\Romannum{2}}. When plugging Eqn.\ref{Eq12} and Eqn.\ref{Eq13} into Eqn.\ref{Eq11} to achieve Eqn.\ref{Eq15}, we employ the properties of matrix inner product $\bm{A}:\bm{B} = Tr(\bm{A}^T\bm{B})$, which were studied in \cite{ionescu2016training}, to derive the following equivalent equation
\begin{alignat}{1}
	\bm{A}:\bm{B}_{asym} = \bm{A}_{bsym}:\bm{B},  \label{Eq29}
\end{alignat}
where $\bm{B}_{asym}=\bm{B}_{tril}-(\bm{B}_{tril})^T$, $\bm{A}_{bsym}=\bm{A}_{tril}-(\bm{A}^T)_{tril}$, $\bm{A}_{tril}$ extracts the elements below the main diagonal of $\bm{A}$.


\end{document}